\documentclass[twocolumn]{article}

\usepackage{arxiv}

\usepackage{multicol}
\usepackage[utf8]{inputenc} % allow utf-8 input
\usepackage[T1]{fontenc}    % use 8-bit T1 fonts
\usepackage{hyperref}       % hyperlinks
\usepackage{url}            % simple URL typesetting
\usepackage{booktabs}       % professional-quality tables
\usepackage{amsfonts}       % blackboard math symbols
\usepackage{nicefrac}       % compact symbols for 1/2, etc.
\usepackage{microtype}      % microtypography
\usepackage{graphicx}
\usepackage{doi}

\usepackage[
    backend=biber,
    bibstyle=ieee,
    citestyle=numeric-comp,
    hyperref=true,
    date=year,
    bibencoding=inputenc,
    uniquename=init,
    giveninits=true
]{biblatex}
\usepackage{longtable}
\usepackage{multirow} 
\usepackage{amsmath}
\usepackage{siunitx}
\usepackage{adjustbox}
\usepackage{subcaption}
\usepackage{acro}
\usepackage{xspace}
\usepackage{upgreek}
\usepackage{algorithm}
\usepackage{algorithmicx}
\usepackage{algpseudocode}
\usepackage{algcompatible}
\usepackage{relsize}
\usepackage{nth}
\usepackage[shortcuts]{extdash}
\usepackage{pifont}
\usepackage{mathtools}
\usepackage{newtxmath}
\usepackage{amsfonts}

\newcommand\ie{i.\,e.\xspace}
\newcommand\eg{e.\,g.\xspace}
\DeclareMathSymbol{\shortminus}{\mathbin}{AMSa}{"39}

\DeclareAcronym{ANN}{
    short = ANN,
    long  = Artificial Neural Network,
    tag = acronyms
}
\DeclareAcronym{WP}{
    short = WP,
    long  = Wind Power,
    tag = acronyms
}
\DeclareAcronym{PV}{
    short = PV,
    long  = PhotoVoltaic,
    tag = acronyms
}
\DeclareAcronym{ARMA}{
    short = ARMA,
    long  = Autoregressive Moving Average,
    tag = acronyms
}
\DeclareAcronym{ARIMA}{
    short = ARIMA,
    long  = Autoregressive Integrated Moving Average,
    tag = acronyms
}
\DeclareAcronym{ARIMAX}{
    short = ARIMAX,
    long  = Autoregressive Integrated Moving Average with eXternal Input,
    tag = acronyms
}
\DeclareAcronym{LSTM}{
    short = LSTM,
    long  = Long Short-Term Memory,
    tag = acronyms
}
\DeclareAcronym{DT}{
    short = DT,
    long  = Decision Tree,
    tag = acronyms
}
\DeclareAcronym{MLP}{
    short = MLP,
    long  = MultiLayer Perceptron,
    tag = acronyms
}
\DeclareAcronym{RF}{
    short = RF,
    long  = Random Forest,
    tag = acronyms
}
\DeclareAcronym{SVR}{
    short = SVR,
    long  = Support Vector Regression,
    tag = acronyms
}
\DeclareAcronym{XGB}{
    short = XGB,
    long  = eXtreme Gradient Boosting,
    tag = acronyms
}
\DeclareAcronym{DeepAR}{
    short = DeepAR,
    long  = Deep AutoRegression,
    tag = acronyms
}
\DeclareAcronym{NHiTS}{
    short = N-HiTS,
    long  = Neural Hierarchical interpolation for Time Series forecasting,
    tag = acronyms
}
\DeclareAcronym{TFT}{
    short = TFT,
    long  = Temporal Fusion Transformer,
    tag = acronyms
}
\DeclareAcronym{LOF}{
    short = LOF,
    long  = Local Outlier Factor,
    tag = acronyms
}
\DeclareAcronym{ECMWF}{
    short = ECMWF,
    long  = European Centre for Medium-Range Weather Forecasts,
    tag = acronyms
}
\DeclareAcronym{nMAE}{
    short = nMAE,
    long  = normalized Mean Absolute Error,
    tag = acronyms
}
\DeclareAcronym{nRMSE}{
    short = nRMSE,
    long  = normalized Root Mean Squared Error,
    tag = acronyms
}
\DeclareAcronym{OEP}{
    short = OEP,
    long  = Open Energy Platform,
    tag = acronyms
}
\DeclareAcronym{ML}{
    short = ML,
    long  = Machine Learning,
    tag = acronyms
}
\DeclareAcronym{DL}{
    short = DL,
    long  = Deep Learning,
    tag = acronyms
}
\DeclareAcronym{OEM}{
    short = OEM,
    long  = Original Equipment Manufacturer,
    tag = acronyms
}
\DeclareAcronym{SM}{
    short = SM,
    long  = Statistical Modeling,
    tag = acronyms
}
\DeclareAcronym{PK}{
    short = PK,
    long  = Prior Knowledge,
    tag = acronyms
}
\DeclareAcronym{HPO}{
    short = HPO,
    long  = HyperParameter Optimization,
    tag = acronyms
}
\DeclareAcronym{AutoML}{
    short = AutoML,
    long  = Automated Machine Learning,
    tag = acronyms
}

\title{On autoregressive deep learning models for day-ahead wind power forecasting with irregular shutdowns due to redispatching}
\date{} 					% Or removing it

\author{
    \href{https://orcid.org/0000-0002-9320-5341}{\includegraphics[scale=0.06]{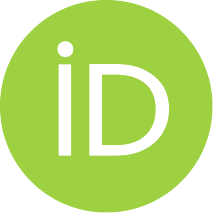}\hspace{1mm}
    Stefan Meisenbacher}\\
	Karlsruhe Institute of Technology\\
	Eggenstein-Leopoldshafen, 76344, Germany\\
	\texttt{stefan.meisenbacher@kit.edu}\\
	\And
    \href{https://orcid.org/0009-0008-5651-9932}{\includegraphics[scale=0.06]{orcid.pdf}\hspace{1mm}
    Silas Aaron Selzer}\\
	University of Wuppertal\\
	42119 Wuppertal, Germany\\
        Technische Universität Ilmenau,\\
	98693 Ilmenau, Germany\\
	\And
    \href{}{\includegraphics[scale=0.06]{orcid.pdf}\hspace{1mm}
    Mehdi Dado}\\
	Karlsruhe Institute of Technology\\
	Eggenstein-Leopoldshafen, 76344, Germany\\
	\And
    \href{https://orcid.org/0009-0009-9713-3590}{\includegraphics[scale=0.06]{orcid.pdf}\hspace{1mm}
    Maximilian Beichter}\\
	Karlsruhe Institute of Technology\\
	Eggenstein-Leopoldshafen, 76344, Germany\\
	\And
    \href{https://orcid.org/0000-0002-1260-179X}{\includegraphics[scale=0.06]{orcid.pdf}\hspace{1mm}
    Tim Martin}\\
	Karlsruhe Institute of Technology\\
	Eggenstein-Leopoldshafen, 76344, Germany\\
	\And
    \href{}{\includegraphics[scale=0.06]{orcid.pdf}\hspace{1mm}
    Markus Zdrallek}\\
	University of Wuppertal\\
	42119 Wuppertal, Germany\\
	\And
    \href{}{\includegraphics[scale=0.06]{orcid.pdf}\hspace{1mm}
    Peter Bretschneider}\\
        Technische Universität Ilmenau,\\
	98693 Ilmenau, Germany\\
        Fraunhofer IOSB,
        98693 Ilmenau, Germany\\
	\And
    \href{https://orcid.org/0000-0002-3572-9083}{\includegraphics[scale=0.06]{orcid.pdf}\hspace{1mm}
    Veit Hagenmeyer}\\
	Institute for Automation and Applied Informatics\\
	Karlsruhe Institute of Technology\\
	Eggenstein-Leopoldshafen, 76344, Germany\\
    \And
    \href{https://orcid.org/0000-0001-9100-5496}{\includegraphics[scale=0.06]{orcid.pdf}\hspace{1mm}
    Ralf Mikut}\\
	Institute for Automation and Applied Informatics\\
	Karlsruhe Institute of Technology\\
	Eggenstein-Leopoldshafen, 76344, Germany\\
}

\renewcommand{\headeright}{Meisenbacher \textit{et al.}}
\renewcommand{\undertitle}{}
\renewcommand{\shorttitle}{Day-ahead wind power forecasting with regular and irregular shutdowns}

\hypersetup{
pdftitle={On autoregressive deep learning models for day-ahead wind power forecasting with irregular shutdowns due to redispatching},
pdfsubject={day-ahead wind power forecasting with regular and irregular shutdowns},
pdfauthor={Meisenbacher et al.},
pdfkeywords={
        deep learning,
        autoregression,
        data cleaning,
        redispatch,
        wind power curve,
        time series forecasting
    },
}

\begin{document}
\twocolumn[
    \begin{@twocolumnfalse}
        \maketitle
        \begin{abstract}
            Renewable energies and their operation are becoming increasingly vital for the stability of electrical power grids since conventional power plants are progressively being displaced, and their contribution to redispatch interventions is thereby diminishing.
            In order to consider renewable energies like \acf{WP} for such interventions as a substitute, day-ahead forecasts are necessary to communicate their availability for redispatch planning.
            In this context, automated and scalable forecasting models are required for the deployment to thousands of locally-distributed onshore \acs{WP} turbines.
            Furthermore, the irregular interventions into the \acs{WP} generation capabilities due to redispatch shutdowns pose challenges in the design and operation of \acs{WP} forecasting models.
            Since state-of-the-art forecasting methods consider past \acs{WP} generation values alongside day-ahead weather forecasts, redispatch shutdowns may impact the forecast.
            Therefore, the present paper highlights these challenges and analyzes state-of-the-art forecasting methods on data sets with both regular and irregular shutdowns.
            Specifically, we compare the forecasting accuracy of three autoregressive \acf{DL} methods to methods based on \acs{WP} curve modeling.
            Interestingly, the latter achieve lower forecasting errors, have fewer requirements for data cleaning during modeling and operation while being computationally more efficient, suggesting their advantages in practical applications.
        \end{abstract}
        % keywords can be removed
        \keywords{
            deep learning,
            autoregression,
            data cleaning,
            redispatch,
            wind power curve,
            time series forecasting
        }
    \end{@twocolumnfalse}
]

\clearpage

\section{Introduction}
\label{sec:introduction}
Forecasting locally distributed \ac{WP} generation is required to prevent grid congestion by balancing the electrical transmission and distribution grids \cite{Salm2022}.
As \ac{WP} capacity expands in future energy systems, automating the design and operation of \ac{WP} forecasting models becomes inevitable to keep pace with \ac{WP} capacity expansion.
However, two key challenges exist in developing such automated \ac{WP} forecasting models:

First, scalable \ac{WP} forecasting models that achieve low forecasting errors are needed.
On the one hand, offshore \ac{WP} farms, consisting of many centralized \ac{WP} turbines in a uniform terrain, lend themselves to holistic modeling approaches.
Such models can incorporate the wind direction to account for wake losses (mutual influence of \ac{WP} turbines).
In holistic modeling, the use of computationally intensive methods can be worthwhile, for example as in \cite{Meka2021} using an autoregressive \ac{DL} model.
On the other hand, onshore \ac{WP} turbines are created decentrally in different terrains, \eg, in areas with little air turbulence like open fields or in the area of cities, forests, and hills with higher air turbulence.
The diversity of onshore \ac{WP} systems, featuring turbines with varying characteristics and located in heterogeneous terrains, prompts questions about the cost\-/effectiveness of employing computationally intensive \ac{DL} models for forecasting.

Second, interventions in the \ac{WP} generation capabilities at regular and irregular intervals represent a challenge for designing and operating \ac{WP} forecasting models.
Regular shutdowns comprise time\-/controlled shutdowns, such as those for bat protection and maintenance, while changing the operating schedule to prevent line overloads (redispatch) represents irregular shutdowns.\footnote{
    Other reasons for irregular shutdowns are sudden faults and the provision of balancing energy for frequency control.
}
For planning redispatch interventions, the forecast is required to communicate the available \ac{WP} generation without incorporating any forecasts of future redispatch interventions.
Hence, the preparation of the training data sub\-/set and model design must take this forecasting target into account.
Despite the availability of various training data cleaning methods \cite{Wang2021, Su2019, Luo2022}, redispatch interventions can also impair the model operation in making forecasts if they rely on autoregression.
More precisely, such methods make a \ac{WP} forecast based on a horizon of past target time series values alongside exogenous forecasts (future covariates) like wind speed and direction forecasts.
Since redispatch\-/related shutdowns also appear in this horizon of past values, the model may also forecast future shutdowns, which has an undesirable impact on forecast\-/based planning \cite{Klaiber2015}.

Therefore, the present paper increases awareness of both challenges and compares autoregressive forecasting methods on data sets containing regular and irregular shutdowns.
Three autoregressive methods are considered, namely \ac{DeepAR} \cite{Salinas2020}, \ac{NHiTS} \cite{Challu2023}, and the \ac{TFT} \cite{Lim2021b}, which are selected for comparison since they represent distinct state\-/of\-/the\-/art \ac{DL} architectures.
These autoregressive forecasting methods are compared with methods based on \ac{WP} curve modeling, which rely solely on weather forecasts and omit past values.
For \ac{WP} curve modeling, the three \ac{ML} methods \ac{XGB}, \ac{SVR} and the \ac{MLP} are considered.
Additionally, the \ac{OEM}'s \ac{WP} curve and an \ac{AutoML} method based on \ac{WP} curves called AutoWP \cite{Meisenbacher2024d} are included.
In contrast to other \ac{AutoML} methods \cite{Meisenbacher2022}, AutoWP dispenses with the computationally expensive \ac{HPO} but uses computationally efficient ensemble learning, making it scalable for model deployment to thousands of individual \ac{WP} turbines.

This article is an extension of the conference paper \cite{Meisenbacher2024d} and is organized as follows: related work is reviewed in \autoref{sec:related_work}, applied methods are detailed in \autoref{sec:methodology}, the evaluation is given in \autoref{sec:evaluation} and results are discussed in \autoref{sec:discussion}, followed by the conclusion and outlook in \autoref{sec:conclusion_outlook}.

\section{Related work}
\label{sec:related_work}
This section first focuses on renewable energy forecasting before \ac{WP} forecasting methods are analyzed in the context of redispatch interventions.\footnote{
    Note that other applications, \eg, model\-/based observer and control design \cite{Schulte2024}, require different modeling methods.
}
Finally, related work that considers inconsistent \ac{WP} data is reviewed and the research gap is summarized.

As stated in the introduction, the growing share of renewable power generation results in an increasing demand for forecasts.
In this context, \ac{PV} \cite{Meisenbacher2023a, Sweeney2020, LopezSantos2022} and \ac{WP} forecasts \cite{Meisenbacher2024d, Sweeney2020, vanHeerden2023} are strongly weather\-/dependent.
For \ac{WP} forecasts, the methods can be separated into two types:
\begin{enumerate}
    \item The first type comprises autoregressive methods, \ie, the forecast relies on the target turbine's past \ac{WP} generation values.
    Weather forecasts can be incorporated as exogenous features.
    \item The second type comprises methods that do not consider past values but instead model the so\-/called \ac{WP} curve, \ie, the empirical relationship between wind speed and \ac{WP}.
    Further additional explanatory variables can be taken into account.
\end{enumerate}

With regard to autoregressive methods (first type), methods based on \ac{SM}, \ac{ML}, and \ac{DL} are proposed in the literature.
Classical autoregressive \ac{SM} methods include the \ac{ARMA} \cite{Rajagopalan2009} and its integrated version, the \acs{ARIMA} \cite{Hodge2011}, to cope with trends.
Whilst they disregard the strong weather\-/dependency of \ac{WP} generation, it can be considered by extending \ac{SM} with exogenous variables, \eg \acs{ARIMAX} \cite{Ahn2023}.
Exogenous variables can also be considered in \ac{ML}\-/based autoregressive models by considering both exogenous variables and past values as features in the regression method, \eg, in \ac{DT}\-/based methods like \ac{XGB} \cite{Sobolewski2023} or the \ac{MLP} \cite{Lahouar2017}.
Apart from the weather, additional information can be taken into account.
For example, the authors in \cite{Sobolewski2023} encode seasonal relationships by cyclic features and compare regression methods to the \ac{LSTM}, a special deep \ac{ANN} architecture.\footnote{
    Deep \acp{ANN} with special architectures belong to \ac{DL}, which is a sub\-/set of \ac{ML}.
}
The literature on \ac{DL}-based \ac{WP} forecasting comprises architectures consisting of convolutional layers, recurrent and residual connections \cite{Liu2019, Zhou2019, Zhu2020, Shahid2021, Arora2023}, as well as transformer architectures based on the attention\-/mechanism \cite{Sun2023, vanHeerden2023, Hu2024, Xu2024, Mo2024}.
Although methods based on \ac{DL} can learn complex covariate and temporal relationships, their scalability is limited due to the high computational training effort.
Moreover, all autoregressive methods are subject to concerns regarding shutdowns due to redispatch interventions.
In fact, methods based on autoregression can be vulnerable to forecast redispatch interventions, which is an undesirable outcome for redispatch planning.

With regard to methods for \ac{WP} curve modeling (second type), parametric and non\-/parametric approaches are proposed in the literature \cite{Lydia2014}.
Parametric modeling approaches are based on fitting assumed mathematical expressions, such as polynomial, cubic, and exponential expressions \cite{Carrillo2013}.
Specifically, the \ac{WP} curve is commonly separated into four sections, separated by the cut\-/in, nominal, and cut\-/out wind speed \cite{Lydia2013}.
Note that such \ac{WP} curves only apply to the wind speed at the height where the empirical measurements originate from.
Non\-/parametric approaches comprise \ac{ML}\-/based methods, \eg, \ac{DT}\-/based methods \cite{Pandit2019, Moreno2020, Singh2021}, \ac{SVR} \cite{Pandit2019, Wang2019b, Moreno2020}, and the \ac{MLP} \cite{Li2001, Schlechtingen2013, Wang2019b}.
Further, \ac{ML}\-/based methods can also consider additional explanatory variables such as air temperature \cite{Schlechtingen2013, RodriguezLopez2020}, atmospheric pressure \cite{RodriguezLopez2020}, and wind direction \cite{Li2001, Schlechtingen2013, Singh2021}.
However, inconsistent data is a crucial challenge for \ac{WP} curve modeling, as detailed below.

Inconsistent data due to shutdowns or partial load operation represent a challenge for the modeling of normal \ac{WP} turbine operation.
According to \cite{Morrison2022}, abnormal operation manifests in three principal types: power generation by stop\-/to\-/operation transitions and vice versa, steady power generation at a power less than the turbine's peak power rating, and no power generation whilst above cut\-/in wind speed.
To address this, data cleaning methods are commonly employed, \eg, removing power generation samples that deviate from an already fitted \ac{WP} curve \cite{Wang2019b} or are lower than a threshold near zero \cite{Schlechtingen2013, Wang2019b}; additionally, experts can label abnormal operation \cite{Moreno2020}.
Expert knowledge is not required for automated outlier detection methods like the sample-wise calculation of a \ac{LOF}, where samples having a significantly lower \ac{LOF} than their neighbors are labeled as outliers \cite{Morrison2022}.
Although pre\-/processing of inconsistent data is commonly used in \ac{WP} curve modeling, it still remains unused in many state\-/of\-/the\-/art methods based on autoregression \cite{Zhu2020, Shahid2021, Sobolewski2023, vanHeerden2023, Xu2024, Mo2024}.
Even when applied, it is limited to training data cleaning \cite{Wang2019b, Hu2024}.
However, it is unexplored how inconsistent \ac{WP} turbine operation data impact the error of forecasting models based on autoregression, \ie, models that consider a horizon of past power generation values to compute the forecast.

\section{Wind power forecasting}
\label{sec:methodology}

This section introduces considered autoregressive \ac{DL}-based methods and \ac{WP} curve modeling-based methods for \ac{WP} forecasting.
It further explores the application of \ac{PK} for post\-/processing model outputs, along with data processing methods for handling shutdowns in both model design and model operation.

\subsection{Forecasting methods based on autoregression and deep learning}

A forecasting model based on autoregression $f\left(\cdot\right)$ estimates future expected values $\hat{y}$ at the origin $k$ for the forecast horizon $H \in \mathbb{N}_1$ using past and current values \cite{GonzalezOrdiano2018}.
Formally, such a model is defined as
\begin{equation}
    \begin{split}
    \hat{\mathbf{y}}\left[k \!+\! 1, \ldots, k \!+\! H\right] = f\left(\right.
                        & \mathbf{y}\left[k \!\shortminus\! H_1, \ldots, k\right], \\
                        & \mathbf{X}\left[k \!\shortminus\! H_1, \ldots, k\right], \\
                        & \hat{\mathbf{X}}\left[k \!+\! 1, \ldots, k \!+\! H\right], \\
                        & \left.\mathbf{p}\right), k, H, H_1 \in \mathbb{N}_1, k > H_1,
    \end{split}
    \label{eq:intro_point-forecasting-model}
\end{equation}
where $\mathbf{y}$ represents values of the \ac{WP} turbine's power generation, the matrix $\mathbf{X}$ represents covariates, $\hat{\mathbf{X}}$ indicates that the respective covariates are estimates, $H_1 \in \mathbb{N}_1$ is the horizon of past values $k \!\shortminus\! H_1$, and the vector $\mathbf{p}$ includes the model's trainable parameters~\cite{GonzalezOrdiano2018}.

For $\mathbf{X}$, further explanatory variables for shutdown handling are considered that are detailed in \autoref{ssec:autowp_shutdown-handling};
for $\hat{\mathbf{X}}$, the exogenous forecasts of air temperature and atmospheric pressure, as well as wind speed and direction are taken into account;
and for $H_1$, past values up to one day are considered, \ie, $96$ values since the time resolution is 15 minutes.

State\-/of\-/the\-/art forecasting methods based on autoregression often use \ac{DL}, \ie, (deep) \acp{ANN} with special architectures, such as recurrent and residual connections, as well as attention units.
We consider the three \ac{DL} methods \ac{DeepAR} \cite{Salinas2020}, \ac{NHiTS} \cite{Challu2023}, and \ac{TFT} \cite{Lim2021b}, and utilize the implementations provided by the Python programming framework PyTorch Forecasting \cite{Beitner2020}.

\paragraph*{\textbf{DeepAR}}
The forecasting method \cite{Salinas2020} leverages recurrent \ac{LSTM} layers \cite{Hochreiter1997} to capture temporal dependencies in the data.
Recurrent layers allow -- unlike uni-directional feed-forward layers -- the output from neurons to affect the subsequent input to the same neuron.
Such a neuron maintains a hidden state that captures information about the past observations in the time series, i.e., the output depends on the prior elements within the input sequence.
Consequently, \ac{DeepAR} is autoregressive in the sense that it receives the last observation as input, and the previous output of the network is fed back as input for the next time step.

\paragraph*{\textbf{N-HiTS}}
The forecasting method \cite{Challu2023} is based on uni-directional feed-forward layers with ReLU activation functions.
The layers are structured as blocks, connected using the double residual stacking principle.
Each block contains two residual branches, one makes a backcast and the other makes a forecast.
The backcast is subtracted from the input sequence before entering the next block, effectively removing the portion of the signal that is already approximated well by the previous block.
This approach simplifies the forecasting task for downstream blocks.
The final forecast is produced by hierarchically aggregating the forecasts from all blocks.

\paragraph*{\textbf{TFT}}
The forecasting method \cite{Lim2021b} combines recurrent layers with the transformer architecture \cite{Vaswani2017}, which is built on attention units.
An attention unit computes importance scores for each element in the input sequence, allowing the model to focus on those elements in a sequence that significantly influence the forecast.
While the attention units are used to capture long-term dependencies, short-term dependencies are captured using recurrent layers based on the \ac{LSTM}.

\subsection{Forecasting methods based on wind power curve modeling}

\ac{WP} curve modeling seeks to establish a temporally\-/independent empirical relationship between wind speed, \ac{WP}, and possibly, additional explanatory variables.
In the following, the \ac{WP} curve and its application in \ac{WP} forecasting are introduced.
Afterward, AutoWP is presented before \ac{ML} methods for \ac{WP} curve modeling are outlined.

\paragraph*{Wind power curve}

\begin{figure}
    \begin{subfigure}[t]{0.49\columnwidth}
        % \tikzset{external/remake next}
        \includegraphics{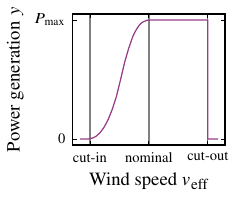}
        \caption{%\mathversion{lm}
            The \acs{WP} curve is the power generation as a function of the wind speed at a reference height and consists of four sections separated by the cut\-/in, nominal, and cut\-/out wind speed.
        }
        \label{fig:autowp_wind-power-curve_line-plot}
    \end{subfigure}
    \hfill
    \begin{subfigure}[t]{0.49\columnwidth}
        \includegraphics{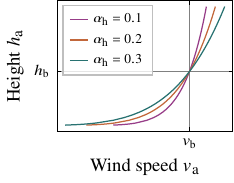}
        \caption{%\mathversion{lm}
            Height correction using the wind profile power law \eqref{eq:autowp_wind-power-law} with the known wind speed $v_\text{b}$ at height $h_\text{b}$ above ground level outlined for different values of the exponent $\alpha_\text{h}$.
        }
        \label{fig:autowp_wind-height-correction_line-plot}
    \end{subfigure}
    \caption{%\mathversion{lm}
        Using the \acs{WP} curve to make forecasts requires the wind speed forecast to be corrected to the curve's reference height.
    }
    \label{fig:autowp_wind-power-curve_multi-plot}
\end{figure}

A \ac{WP} curve $y\left[k\right] = f\left(v_\text{eff}\left[k\right]\right)$ describes the turbine's power generation $y\left[k\right]$ as a function of the wind speed $v_\text{eff}\left[k\right]$ at a reference height above ground level.
The curve is separated into four sections by the cut\-/in, nominal, and cut\-/out wind speed, see \autoref{fig:autowp_wind-power-curve_line-plot}.
The reference height of \ac{WP} curves provided by turbine \acp{OEM} is commonly the hub height.
Using wind speed forecasts as inputs for \ac{OEM} \ac{WP} curves necessitates a height correction,\footnote{
    Unless the turbine coincidentally has a hub height of $\SI{100}{\meter}$.
}
as wind speed forecasts are commonly available at $\SI{10}{\meter}$ and $\SI{100}{\meter}$ above ground level.
Differences in wind speed at different heights arise from the interaction between the wind and the earth's surface (so\-/called terrain roughness) and the wind (so\-/called atmospheric boundary layer) \cite{Etling2008}.
A common and straightforward modeling approach for this phenomenon is the wind power law \cite{Jung2021}
\begin{equation}
    \frac{v_\text{a}}{v_\text{b}} = \left(\frac{h_\text{a}}{h_\text{b}}\right)^{\alpha_\text{h}},
    \label{eq:autowp_wind-power-law}
\end{equation}
where the empirically\-/determined exponent $\alpha_\text{h}$ depends on the terrain \cite{Masters2004} and is used to establish the relation between the wind speeds $v_\text{a}$ and $v_\text{b}$ at the heights $h_\text{a}$ and $h_\text{b}$ above ground level.
The impact of different values of $\alpha_\text{h}$ is exemplified in \autoref{fig:autowp_wind-height-correction_line-plot}.
For the height correction of the wind speed forecast based on \eqref{eq:autowp_wind-power-law}, we perform two steps \cite{Meisenbacher2021a}:
First, we take $\alpha_\text{h} = 1/9$ for offshore and $\alpha_\text{h} = 1/7$ for onshore \ac{WP} turbines \cite{Hsu1994}, and estimate the turbine's effective hub height
\begin{equation}
    h_\text{eff} = \SI{100}{\meter} \cdot \left(\frac{\overline{v}_\text{eff}}{\overline{\hat{v}}_{100}}\right)^{(1/\alpha_\text{h})}
    \label{eq:autowp_effective-hub-height}
\end{equation}
with the averages of the wind speed forecast at $\SI{100}{\meter}$ $\overline{\hat{v}}_{100}$ and the wind speed measurement at hub height $\overline{v}_\text{eff}$ on the training data sub\-/set.
Second, each value of the wind speed forecast $\hat{v}_{100}[k]$ can now be corrected to hub height using
\begin{equation}
    \hat{v}_\text{eff}[k] = \hat{v}_{100}[k] \cdot \left(\frac{h_\text{eff}}{\SI{100}{\meter}}\right)^{\alpha_\text{h}}.
    \label{eq:autowp_effective-wind-speed}
\end{equation}
The advantage of this two\-/step approach is that knowledge about the actual hub height is not required and extrapolations from the $\SI{100}{\meter}$ reference are near the hub height of today's \ac{WP} turbines with a typical hub height between $\SI{80}{\meter}$ and $\SI{140}{\meter}$ above ground level \cite{Jung2021}.
Finally, the height-corrected wind speed forecast $\hat{v}_\text{eff}[k]$ is used as the \ac{WP} curve's input to forecast the turbine's power generation:
\begin{equation}
    \hat{y}[k] = f\left(\hat{v}_\text{eff}[k]\right).
    \label{eq:autowp_wp-curve-forecast}
\end{equation}

\paragraph*{AutoWP}

\begin{figure*}
    \centering
    \begin{subfigure}[t]{.32\textwidth}
        \centering
        \includegraphics{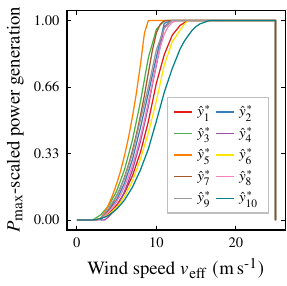}
        \caption{%\mathversion{lm}
            The first step creates the ensemble pool using $N_\text{m}=10$ normalized \acs{WP} curves $\hat{y}^{*}$, $n \in \mathbb{N}_1^{N_\text{m}}$ of the \acs{OEP} wind turbine library \cite{Petersen2019}.
            The selection reduces redundancy and preserves diversity.
        }
        \label{fig:autowp_method-plot-1_wka3_line-plot}
    \end{subfigure}%
    \hfill
    \begin{subfigure}[t]{.32\textwidth}
        \centering
        \includegraphics{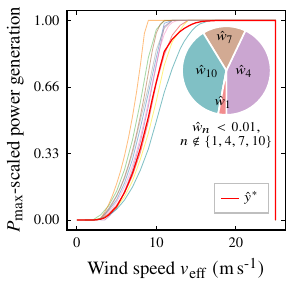}
	\caption{%\mathversion{lm}
            The seconds step computes the normalized ensemble \acs{WP} curve $\hat{y}^{*}$ as convex linear combination of the pool's curves, with the weights $\hat{w}_{n}$, $n \in \mathbb{N}_1^{N_\text{m}}$ adapted to optimally fit the new \acs{WP} turbine.
        }
	\label{fig:autowp_method-plot-2_wka3_line-plot}
	\end{subfigure}%
    \hfill
    \begin{subfigure}[t]{.32\textwidth}
        \centering
        \includegraphics{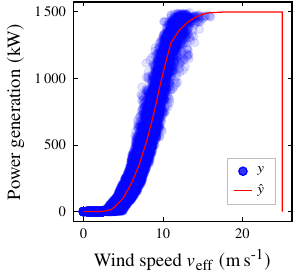}
	\caption{%\mathversion{lm}
            The third step re\-/scales the ensemble \acs{WP} curve $\hat{y}$ to the peak power rating $P_\text{max,new} = \SI{1500}{\kilo\watt}\text{p}$ of the new \acs{WP} turbine.
            Additionally, the measurement $y$ used to fit the ensemble weights is shown.
        }
	\label{fig:autowp_method-plot-3_wka3_line-plot}
	\end{subfigure}
    \caption{%\mathversion{lm}
        The three steps of AutoWP's automated design exemplified for the real\-/world \acs{WP} turbine no. 1.
    }
    \label{fig:autowp_method-plot_multi-plot}
\end{figure*}

AutoWP is based on the underlying idea of representing a new \ac{WP} turbine by the optimally weighted sum of \ac{WP} curves a sufficiently diverse ensemble.\footnote{
    Repository: \url{https://github.com/SMEISEN/AutoWP}
}
The method includes three steps \cite{Meisenbacher2024d}: i) creation of the ensemble with normalized \ac{WP} curves, ii) computation of the normalized ensemble \ac{WP} curve using the optimally weighted sum of considered normalized \ac{WP} curves, and iii) re\-/scaling of the ensemble \ac{WP} curve based on the peak power rating of the new \ac{WP} turbine.
These three steps are exemplified in \autoref{fig:autowp_method-plot_multi-plot} and detailed in the following.

In the first step, the ensemble is created using \ac{OEM} \ac{WP} curves from the wind turbine library of the \ac{OEP} \cite{Petersen2019}.
This process involves loading the database, selecting all entries that include a \ac{WP} curve, resampling the $\{P_n, v_\text{eff}\}$ value pairs to ensure a uniform wind speed sampling rate, and normalizing the power values $P_n[v_\text{eff}]$ based on the peak power rating of each \ac{WP} curve $P_{\text{max},n}$:
\begin{equation}
    P_n^*[v] = \frac{P_n[v_\text{eff}]}{P_{\text{max},n}}.
    \label{eq:autowp_scaling-curves}
\end{equation}
Such a normalized \ac{WP} curve outputs $\hat{y}^*[k] = f\left(v_\text{eff}[k]\right)$ as a function of the wind speed $v_\text{eff}[k], k \in \mathbb{N}^K$.
Limiting the size of the ensemble without losing diversity is achieved by sorting all \ac{WP} curves according to their sum of normalized power values, keeping the first and last \ac{WP} curve, together with eight \ac{WP} curves in between at equally spaced intervals, \ie, $N_\text{m}=10$.

In the second step, we compute the normalized ensemble \ac{WP} curve using the weighted sum of the considered \ac{WP} curves
\begin{equation}
    \hat{y}^*[k] = \sum^{N_\text{m}}_{n=1} \hat{w}_n \cdot \hat{y}_n^*[k],
    \label{eq:autowp_weighted-sum}
\end{equation}
which is a convex linear combination with $\hat{w}_n$, $n \in \mathbb{N}_1^{N_\text{m}}$ and $\hat{y}_n$ being the weight and output of the $n$-th curve in the ensemble.
In order to determine the optimal weights, the target time series $y$ is normalized analogously to \eqref{eq:autowp_scaling-curves}
\begin{equation}
    y^*[k] = \frac{y[k]}{P_{\text{max},n}},
    \label{eq:autowp_scaling-samples}
\end{equation}
and the optimization problem
\begin{equation}
    \begin{split}
        \min_{\hat{\mathbf{w}}} \frac{1}{K} \sum^K_{k=1} \left(\hat{y}^*[k] - y^*[k] \right)^2\\
        \textrm{s.t.} \quad \hat{\mathbf{w}} \in [0,1], \sum^{N_\text{m}}_{n=1} \hat{w}_n = 1,
        \label{eq:autowp_ensemble-optimization}
    \end{split}
\end{equation}
is solved with the least squares algorithm of the Python package SciPy \cite{Virtanen2020}, and the weights being normalized to hold the constraints.\footnote{
    Alternatively, the constraints could be relaxed by removing the requirement for the weights to sum to one and allowing weights to exceed one.
    This adjustment eliminates the need for the third step, making it suitable for cases where the \ac{WP} turbine's peak power rating is unknown or if the turbine operates at reduced power.
}

In the third step, the ensemble output \eqref{eq:autowp_weighted-sum} is re\-/scaled with the new \ac{WP} turbine's peak power rating $P_\text{max,new}$:
\begin{equation}
    \hat{y}[k] = \hat{y}^*[k] \cdot P_\text{max,new}.
    \label{eq:autowp_re-scaling}
\end{equation}

\paragraph*{Machine learning for wind power curve modeling}
\ac{ML} methods can be used to directly learn the relationship between the wind speed forecast at $\SI{100}{\meter}$ and the \ac{WP} turbine's power generation, eliminating the need for a height\-/correction of the wind speed forecast like \ac{OEM} \ac{WP} curves.
Given the non\-/linear relationship between wind speed and power generation (see \autoref{fig:autowp_wind-power-curve_line-plot}), only non\-/linear \ac{ML} methods are considered in the following.
Established regression methods that perform well in a variety of tasks include \ac{DT}\-/based methods like \ac{XGB}, \ac{SVR}, and the \ac{MLP}.
For considering \ac{WP} curve modeling as a regression problem
\begin{equation}
    \hat{y}\left[k\right] = f\left(\hat{\mathbf{X}}\left[k\right],\mathbf{p}\right),
    \label{eq:autowp_regression-problem}
\end{equation}
the explanatory variables $\hat{\mathbf{X}}$ are the same as for the autoregressive \ac{DL} forecasting methods, namely exogenous forecasts of air temperature and atmospheric pressure, as well as wind speed and direction, and the model's parameters $\mathbf{p}$ are determined by training.
That is, the regression model $f\left(\cdot\right)$ uses the exogenous forecasts $\hat{\mathbf{X}}\left[k\right]$ to estimate the \ac{WP} generation $\hat{y}\left[k\right]$.
Unlike autoregressive forecasting methods, the regression model \eqref{eq:autowp_regression-problem} is static, meaning it does not consider past \ac{WP} generation values when making a forecast.

\subsection{Post-processing}
\ac{PK} about the \ac{WP} curve are used to consider three restrictions:
First, the \ac{WP} turbine's peak power rating limits the power generation.
Second, it is impossible to generate negative \ac{WP}.
Third, if the wind speed $\hat{v}_\text{eff}$ exceeds the cut\-/out speed $v_\text{cut-out}$, the \ac{WP} turbine is shut down.
These three rules
\begin{equation}
    \hat{y}[k] =
        \begin{cases*}
            \hat{y}[k],             & if $\hat{y}[k] > 0$\\
                                    & and $\hat{y}[k] < P_\text{max}$\\
                                    & and $\hat{v}_\text{eff}[k] < v_\text{cut-out}$,\\
            P_\text{max},           & if $\hat{y}[k] > P_\text{max}$\\
                                    & and $\hat{v}_\text{eff}[k] < v_\text{cut-out}$,\\
            0,                      & otherwise,\\
        \end{cases*}
    \label{eq:autowp_post-processing}
\end{equation}
are used for the post\-/processing of model outputs, namely \ac{DeepAR}, \ac{NHiTS}, and \ac{TFT} (autoregressive \ac{DL} methods), and \ac{MLP}, \ac{SVR}, and \ac{XGB} (\ac{ML} methods) in \ac{WP} curve modeling.\footnote{
    Regarding the evaluation data set of this paper, the wind speed measurements at hub height $v_\text{eff}$ are below the cut\-/out wind speed $v_\text{cut-out}$.
}

\subsection{Shutdown handling}
\label{ssec:autowp_shutdown-handling}

\begin{figure}
    % \centering
    \includegraphics{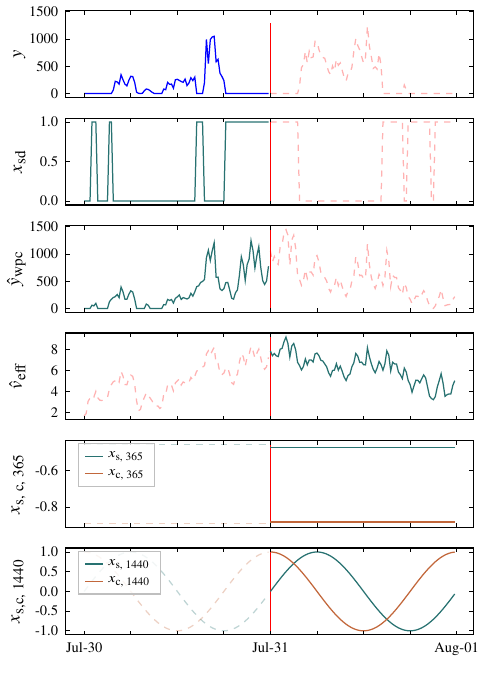}
    \vspace{-0.5mm}
    \caption{%\mathversion{lm}
        Power generation $y$ and covariates available in the forecasting model's future and past horizon.
        $x_\text{sd}$ labels identified shutdowns in the past, $\hat{y}_\text{wpc}$ is the turbine's theoretical power generation according to the \ac{OEM} \ac{WP} curve, $\hat{v}_\text{eff}$ is the wind speed forecast at hub height, and the features $x_\text{s, 365}$, $x_\text{c, 365}$, $x_\text{s, 1440}$, and $x_\text{c, 1440}$ to encode temporally recurring shutdowns patterns.
        Dashed lines visualize unavailable future or unused past periods, and the horizontal line marks the forecast origin.
        }
    \label{fig:autowp_shutdown-handling_explanatory_multi-plot}

    \vspace{1.7mm}

    \hspace{1.1mm}\includegraphics{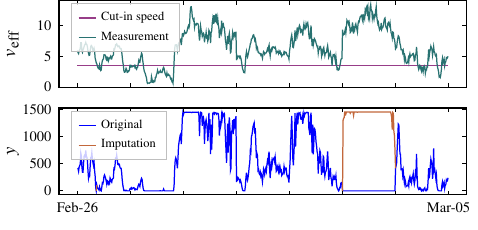}
    \vspace{-0.5mm}
    \caption{%\mathversion{lm}
        Data imputation at shutdowns with the theoretical power generation according to the turbine's \acs{WP} curve (\ie the imputed values depend on the wind speed at hub height $v_\text{eff}$).
        During shutdowns, generation $y$ is zero although the wind speed \(v_\text{eff}\) is higher than cut\-/in speed $v_\text{eff,cut-in}$.
    }
    \label{fig:autowp_shutdown-handling_imputation_multi-plot}
\end{figure}

In the present paper, different shutdown handling methods are taken into account for the model design and the model operation.
For all autoregressive methods, the shutdown handling methods named \textit{none}, \textit{explanatory variables}, \textit{drop}, \textit{imputation}, and \textit{drop\-/imputation} are considered.
For all \ac{WP} curve modeling methods, only the shutdown handling method \textit{drop} is used, as the resulting models operate without a temporal context.

\textit{None} handling of shutdowns refers to the use of raw data for model training and model operation.

\textit{Explanatory variables} refer to the use of additional features.
We use cyclic features as in \cite{Sobolewski2023} to encode temporally recurring shutdowns patterns, namely the day of year and the minute of the day using periodical sine\-/cosine encoding
\begin{align}
x_\text{s, 365}[k]     &= \sin \left(\frac{2 \pi \cdot \text{day}[k]}{365}\right),\\
x_\text{c, 365}[k]     &= \cos \left(\frac{2 \pi \cdot \text{day}[k]}{365}\right)\label{eq:trig_365},\\
x_\text{s, 1440}[k]   &= \sin \left(\frac{2 \pi \cdot \text{minute}[k]}{1440}\right),\\
x_\text{c, 1440}[k]   &= \cos \left(\frac{2 \pi \cdot \text{minute}[k]}{1440}\right).\label{eq:trig_1440}
\end{align}
to establish similarities between related observations, all known in the model's future horizon.\footnote{
    The sine\-/cosine pair is necessary because, otherwise, the encoding would be ambiguous.
}

Moreover, we incorporate additional features only available in the past horizon of the autoregressive methods.
Specifically, these features include the turbine's theoretical power generation according to the \ac{OEM} \ac{WP} curve $\hat{y}_\text{wpc}$, and a categorical feature $x_\text{sd}$ to label samples in the past horizon as normal or abnormal using a rule\-/based and an outlier detection approach.
The former method labels all data points with a wind speed measurement greater than the cut\-/in speed and a power generation lower than the cut\-/in power as shutdowns (similar to \cite{Wang2019b}).
The latter method is based on the \ac{LOF} algorithm (similar to \cite{Morrison2022}) used to identify outliers occurring in \ac{WP} turbine stop\-/to\-/operation transitions and vice versa.
The above\-/introduced additional variables are exemplified in
\autoref{fig:autowp_shutdown-handling_explanatory_multi-plot}.\footnote{
    Since turbine no. 2 experiences shutdowns during nighttime, we introduce an additional feature that is set to 1 during the night and 0 during the day.
}

The shutdown handling method \textit{drop} refers to dropping samples from the training data sub\-/set that are identified as abnormal operational states; and in the shutdown handling method \textit{imputation}, we replace these samples with the turbine's theoretical power generation according to the \ac{OEM} \ac{WP} curve (similar to \cite{Wang2019b}), see \autoref{fig:autowp_shutdown-handling_imputation_multi-plot}.
Finally, we combine the above principle in the shutdown handling method \textit{drop\-/imputation} by identifying abnormal operational states, dropping these samples from the training data sub\-/set, and replacing them during model operation.
In the operation, we only apply the data imputation to data that is available at the forecast origin, \ie, the autoregressive models' past horizon.
The calculation of the assessment metrics is always based on forecasted values and non\-/imputed measurements (ground truth).

\section{Evaluation}
\label{sec:evaluation}

In the evaluation, state\-/of\-/the\-/art forecasting methods based on autoregression are compared with methods that rely solely on day\-/ahead weather forecasts without considering past values.
First, the experimental setup for assessing the forecasting error is described.
Afterward, the results are shown, and insights are provided.

\subsection{Experimental setup}
\label{ssec:autowp_experimental-setup}

The experimental setup introduces the data used for evaluation and the applied evaluation strategy.

\begin{figure}[h!]
    \centering
    \includegraphics{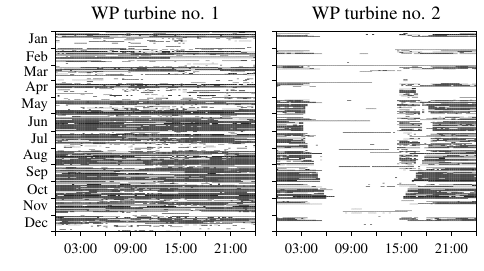}
    \caption{%\mathversion{lm}
        Identified \acs{WP} turbine shutdowns with rule\-/based filtering \cite{Meisenbacher2024d}.
        Black fields represent data points at a measured wind speed greater than the cut\-/in speed with a power generation lower than the cut\-/in power.
        The shutdowns for \acs{WP} turbine no. 1 amount to $\SI{49}{\percent}$ and $\SI{20}{\percent}$ for no. 2.
    }
    \label{fig:autowp_manual-shutdowns_heat-plot}
\end{figure}

\begin{figure}
    \centering
    \includegraphics{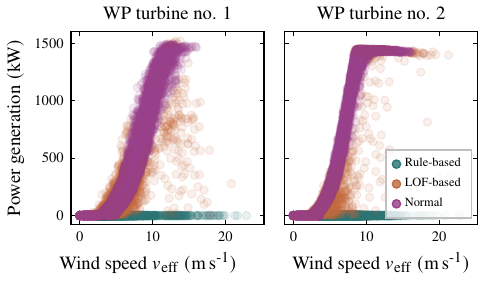}
    \caption{%\mathversion{lm}
        Data pre\-/processing to identify abnormal operational states \cite{Meisenbacher2024d}.
        Rule\-/based filtering identifies turbine shutdowns and \acs{LOF}\-/based filtering identifies turbine stop\-/to\-/operation transitions and vice versa.
        The samples for which the wind speed at hub height is below the cut\-/in speed $v_\text{cut-in}=\SI{2.5}{\meter\per\second}$ amount to $\SI{16.7}{\percent}$, while the cut\-/out speed $v_\text{cut-out}=\SI{25}{\meter\per\second}$ at which the \ac{WP} turbine would be shut down for safety reasons is not reached.
    }
    \label{fig:autowp_abnormal-operation-detection_multi-plot}
\end{figure}

\paragraph*{\textbf{Data}}
For the evaluation, we use a real\-/world data set consisting of two years (2019, 2020) quarter\-/hourly energy metering ($\si{\kWh}$) from two \ac{WP} turbines; \ie, the time resolution is 15 minutes.\footnote{
    Effects with a dynamic below the 15-minute resolution such as stop\-/to\-/operation transitions and vice versa, as well as wind gusts, cannot be modeled.
}
To increase interpretability, we transform the energy metering into the mean power generation time series
\begin{equation}
    \overline{P}[k] = \frac{\Delta E[k]}{t_k},
    \label{eq:autowp_energy-power-transformation}
\end{equation}
with the energy generation $\Delta E[k]$ ($\si{\kilo\watt{}\hour}$) metered within the sample period $t_k$ ($\si{\hour}$).
Additionally, wind speed measurements at turbines' hub heights are available (15-minute average).
Both \ac{WP} turbines have a peak power rating of $\SI{1500}{\kilo\watt}\text{p}$, are located in southern Germany, and are subject to many shutdowns, see \autoref{fig:autowp_manual-shutdowns_heat-plot} and \autoref{fig:autowp_abnormal-operation-detection_multi-plot}, which are not explicitly labeled by the data provider.\footnote{
    Note that the shutdowns are not redispatch interventions since the data set predates the introduction of Redispatch 2.0, which replaces the feed\-/in regulation of the German law EEG and KWKG from 2021 onwards.
}

\noindent

To perform day\-/ahead \ac{WP} forecasting using the methods described in \autoref{sec:methodology}, we use day\-/ahead weather forecasts from the \ac{ECMWF} \cite{ECMWF}.
Importantly, shutdown handling in the autoregressive models' past horizon $H_{1}$ is performed with the weather measurements available up to the forecast origin.

We split the data into a training (2019) and a test data sub\-/set (2020).
Additionally, $\SI{20}{\percent}$ of the training data sub\-/set is hold\-/out for the forecasting methods \ac{DeepAR}, \ac{NHiTS}, and \ac{TFT} to terminate training when the loss on the hold\-/out data increases (early stopping).
% While the wind speed measurement at hub height is used to identify abnormal operation (\autoref{fig:autowp_manual-shutdowns_heat-plot}), the wind speed forecasts are used for model training.

\paragraph*{\textbf{Evaluation strategy}}
The forecasting error is evaluated under two scenarios.
The first scenario concerns day\-/ahead \ac{WP} forecasting including future shutdowns.
It evaluates whether autoregressive forecasting methods can forecast both \ac{WP} generation \textit{and} future shutdowns.
The second concerns day\-/ahead \ac{WP} forecasts to be used for communicating the \ac{WP} turbine's availability for redispatch planning.
The forecast of future shutdowns must be excluded since times of planned shutdowns are communicated via non\-/availability notifications.
The metrics used for assessment are the \ac{nMAE}
\begin{equation}
    \text{nMAE} = \frac{\sum_{k=1}^K\left|\hat{y}[k]-y[k]\right|}{\sum_{k=1}^K{y[k]}},
    \label{eq:autowp_nmae}
\end{equation}
and the \acf{nRMSE}
\begin{equation}
    \text{nRMSE} = \frac{\sqrt{\frac{1}{K} \sum_{k=1}^K\left(\hat{y}[k] - y[k]\right)^2}}{\frac{1}{K} \sum_{k=1}^K y[k]},
    \label{eq:autowp_nrmse}
\end{equation}
with the forecast $\hat{y}[k]$ and the realized value $y[k]$ at time point $k$.

For all methods, the default hyperparameter configuration of the respective implementation is used.
Additionally, methods employing stochastic training algorithm (\ac{DeepAR}, \ac{NHiTS}, \ac{TFT}, and \ac{MLP}) are run five times.

\subsection{Results}
\label{ssec:results}
In the following, the results of the two evaluation scenarios are visualized and summarized.
First, the results of day\-/ahead \ac{WP} forecasting, including shutdowns, are shown to evaluate whether future shutdowns can be forecasted.
Second, the results of day\-/ahead \ac{WP} forecasting disregarding identified shutdowns are shown to evaluate the quality of forecasts used to communicate the turbine's availability for redispatch interventions.
These results are interpreted and discussed in \autoref{sec:discussion}.

\paragraph*{\textbf{Shutdown handling for methods based on autoregression}}

\begin{table}
    \centering
    \caption{%\mathversion{lm}
        The impact of different shutdown handling methods applied to the autoregressive \acs{DL} forecasting methods \acs{DeepAR}, \acs{NHiTS}, and \acs{TFT} on the test \acs{nMAE} when \textit{considering} shutdowns.
        Note that non-autoregressive methods (\acs{WP} curve modeling, \ac{OEM} \ac{WP} curve, and AutoWP) generally perform worse in this scenario, as they cannot forecast shutdowns.
    }
    \label{tab:autowp_shutdown-handling_consider}
    \resizebox{\columnwidth}{!}{%
        \begin{tabular}{ccl|ccc}
        \toprule
        \textbf{Turbine no.} & \textbf{Error} & \textbf{Shutdown handling} & \textbf{\acs{DeepAR}} & \textbf{\acs{NHiTS}} & \textbf{\acs{TFT}} \\
        \midrule
              &       & None  & 1.25$\scriptscriptstyle\pm$\scriptsize0.08 & \textbf{0.92}$\scriptscriptstyle\pm$\scriptsize0.01 & 0.98$\scriptscriptstyle\pm$\scriptsize0.01 \\
              &       & Explanatory variables & 1.14$\scriptscriptstyle\pm$\scriptsize0.04 & 0.93$\scriptscriptstyle\pm$\scriptsize0.03 & 0.94$\scriptscriptstyle\pm$\scriptsize0.01 \\
        1  & \acs{nMAE}     & Drop  & \textbf{1.05}$\scriptscriptstyle\pm$\scriptsize0.09 & 0.95$\scriptscriptstyle\pm$\scriptsize0.03 & \textbf{0.91}$\scriptscriptstyle\pm$\scriptsize0.01 \\
              &       & Imputation & 2.36$\scriptscriptstyle\pm$\scriptsize0.36 & 2.00$\scriptscriptstyle\pm$\scriptsize0.08 & 1.84$\scriptscriptstyle\pm$\scriptsize0.12 \\
              &       & Drop-imputation & 2.61$\scriptscriptstyle\pm$\scriptsize0.30 & 1.92$\scriptscriptstyle\pm$\scriptsize0.03 & 1.78$\scriptscriptstyle\pm$\scriptsize0.04 \\
        \midrule
              &       & None  & 1.15$\scriptscriptstyle\pm$\scriptsize0.17 & 0.76$\scriptscriptstyle\pm$\scriptsize0.02 & 0.81$\scriptscriptstyle\pm$\scriptsize0.02 \\
              &       & Explanatory variables & 1.24$\scriptscriptstyle\pm$\scriptsize0.04 & \textbf{0.73}$\scriptscriptstyle\pm$\scriptsize0.01 & 0.86$\scriptscriptstyle\pm$\scriptsize0.01 \\
        2  & \acs{nMAE}     & Drop  & 1.02$\scriptscriptstyle\pm$\scriptsize0.06 & 0.75$\scriptscriptstyle\pm$\scriptsize0.01 & 0.81$\scriptscriptstyle\pm$\scriptsize0.02 \\
              &       & Imputation & \textbf{0.92}$\scriptscriptstyle\pm$\scriptsize0.03 & 0.80$\scriptscriptstyle\pm$\scriptsize0.01 & \textbf{0.79}$\scriptscriptstyle\pm$\scriptsize0.01 \\
              &       & Drop-imputation & 1.09$\scriptscriptstyle\pm$\scriptsize0.04 & 0.80$\scriptscriptstyle\pm$\scriptsize0.02 & 0.80$\scriptscriptstyle\pm$\scriptsize0.02 \\
        \midrule\\[-13pt]\midrule
              &       & None  & 2.42$\scriptscriptstyle\pm$\scriptsize0.17 & \textbf{2.08}$\scriptscriptstyle\pm$\scriptsize0.02 & 2.11$\scriptscriptstyle\pm$\scriptsize0.04 \\
              &       & Explanatory variables & 2.16$\scriptscriptstyle\pm$\scriptsize0.09 & 2.09$\scriptscriptstyle\pm$\scriptsize0.05 & 2.05$\scriptscriptstyle\pm$\scriptsize0.04 \\
        1 & \acs{nRMSE}     & Drop  & \textbf{2.12}$\scriptscriptstyle\pm$\scriptsize0.15 & 2.15$\scriptscriptstyle\pm$\scriptsize0.05 & \textbf{1.97}$\scriptscriptstyle\pm$\scriptsize0.02 \\
              &       & Imputation & 3.33$\scriptscriptstyle\pm$\scriptsize0.39 & 3.22$\scriptscriptstyle\pm$\scriptsize0.15 & 2.91$\scriptscriptstyle\pm$\scriptsize0.19 \\
              &       & Drop-imputation & 3.60$\scriptscriptstyle\pm$\scriptsize0.36 & 3.07$\scriptscriptstyle\pm$\scriptsize0.03 & 2.79$\scriptscriptstyle\pm$\scriptsize0.06 \\
        \midrule
              &       & None  & 1.57$\scriptscriptstyle\pm$\scriptsize0.18 & 1.17$\scriptscriptstyle\pm$\scriptsize0.03 & 1.22$\scriptscriptstyle\pm$\scriptsize0.02 \\
              &       & Explanatory variables & 1.72$\scriptscriptstyle\pm$\scriptsize0.04 & \textbf{1.10}$\scriptscriptstyle\pm$\scriptsize0.01 & 1.27$\scriptscriptstyle\pm$\scriptsize0.02 \\
        2 & \acs{nRMSE}     & Drop  & 1.42$\scriptscriptstyle\pm$\scriptsize0.09 & 1.13$\scriptscriptstyle\pm$\scriptsize0.02 & 1.20$\scriptscriptstyle\pm$\scriptsize0.02 \\
              &       & Imputation & \textbf{1.26}$\scriptscriptstyle\pm$\scriptsize0.04 & 1.18$\scriptscriptstyle\pm$\scriptsize0.01 & \textbf{1.15}$\scriptscriptstyle\pm$\scriptsize0.02 \\
              &       & Drop-imputation & 1.47$\scriptscriptstyle\pm$\scriptsize0.06 & 1.16$\scriptscriptstyle\pm$\scriptsize0.02 & 1.16$\scriptscriptstyle\pm$\scriptsize0.03 \\
        \bottomrule
        \end{tabular}%
    }
\end{table}

\autoref{tab:autowp_shutdown-handling_consider} shows the impact of different shutdown handling methods on the forecasting error of state\-/of\-/the\-/art autoregressive \ac{DL} methods (\ac{DeepAR}, \ac{NHiTS}, and \ac{TFT}) when \textit{considering shutdowns}.
With regard to \ac{WP} turbine no. 1, the shutdown handling methods \textit{imputation} and \textit{drop\-/imputation} result in significantly higher forecasting errors than the others.
The reason is that numerous shutdowns in the data set (see \autoref{fig:autowp_manual-shutdowns_heat-plot}) are imputed with \ac{OEM} \ac{WP} curve values (forecasting model's past horizon), resulting in forecasts of \ac{WP} generation where the turbine is actually shut down.
For \ac{WP} turbine no. 2, showing regular and time\-/dependent shutdowns (see \autoref{fig:autowp_manual-shutdowns_heat-plot}), considering additional \textit{explanatory variables} only improves over \textit{none} shutdown handling for \ac{NHiTS}, and results in particularly high errors for \ac{DeepAR}.
% The reason for the lack of improvement with additional explanatory variables is that the previous day's shutdown pattern, appearing in the past horizon, may already be a good predictor for the next day.
For both \ac{WP} turbines, shutdown handling methods highly impact the forecasting error, but no method consistently performs best across \ac{DeepAR}, \ac{NHiTS}, and \ac{TFT}.

\begin{table}
    \centering
    \caption{%\mathversion{lm}
        The impact of different shutdown handling methods applied to the autoregressive \acs{DL} forecasting methods \acs{DeepAR}, \acs{NHiTS}, and \acs{TFT} on the test \acs{nMAE} when \textit{disregarding} shutdowns.
    }
    \label{tab:autowp_shutdown-handling_disregard}
    \resizebox{\columnwidth}{!}{%
        % Table generated by Excel2LaTeX from sheet 'Evaluation 1'
        \begin{tabular}{ccl|ccc}
        \toprule
        \textbf{Turbine no.} & \textbf{Error} & \textbf{Shutdown handling} & \textbf{\acs{DeepAR}} & \textbf{\acs{NHiTS}} & \textbf{\acs{TFT}} \\
        \midrule
              &       & None  & 0.93$\scriptscriptstyle\pm$\scriptsize0.06 & \textbf{0.73}$\scriptscriptstyle\pm$\scriptsize0.00 & 0.76$\scriptscriptstyle\pm$\scriptsize0.01 \\
              &       & Explanatory variables & 0.86$\scriptscriptstyle\pm$\scriptsize0.02 & 0.75$\scriptscriptstyle\pm$\scriptsize0.02 & 0.76$\scriptscriptstyle\pm$\scriptsize0.02 \\
        1  & \acs{nMAE}     & Drop  & \textbf{0.81}$\scriptscriptstyle\pm$\scriptsize0.05 & 0.75$\scriptscriptstyle\pm$\scriptsize0.01 & \textbf{0.74}$\scriptscriptstyle\pm$\scriptsize0.01 \\
              &       & Imputation & 1.03$\scriptscriptstyle\pm$\scriptsize0.13 & 0.88$\scriptscriptstyle\pm$\scriptsize0.04 & 0.84$\scriptscriptstyle\pm$\scriptsize0.05 \\
              &       & Drop-imputation & 1.12$\scriptscriptstyle\pm$\scriptsize0.09 & 0.84$\scriptscriptstyle\pm$\scriptsize0.01 & 0.83$\scriptscriptstyle\pm$\scriptsize0.04 \\
        \midrule
              &       & None  & 1.00$\scriptscriptstyle\pm$\scriptsize0.13 & 0.71$\scriptscriptstyle\pm$\scriptsize0.02 & 0.76$\scriptscriptstyle\pm$\scriptsize0.01 \\
              &       & Explanatory variables & 1.10$\scriptscriptstyle\pm$\scriptsize0.03 & 0.67$\scriptscriptstyle\pm$\scriptsize0.01 & 0.80$\scriptscriptstyle\pm$\scriptsize0.01 \\
        2  & \acs{nMAE}     & Drop  & 0.90$\scriptscriptstyle\pm$\scriptsize0.05 & 0.69$\scriptscriptstyle\pm$\scriptsize0.01 & 0.73$\scriptscriptstyle\pm$\scriptsize0.02 \\
              &       & Imputation & \textbf{0.77}$\scriptscriptstyle\pm$\scriptsize0.02 & 0.67$\scriptscriptstyle\pm$\scriptsize0.01 & \textbf{0.67}$\scriptscriptstyle\pm$\scriptsize0.01 \\
              &       & Drop-imputation & 0.90$\scriptscriptstyle\pm$\scriptsize0.04 & \textbf{0.66}$\scriptscriptstyle\pm$\scriptsize0.02 & 0.68$\scriptscriptstyle\pm$\scriptsize0.02 \\
        \midrule\\[-13pt]\midrule
              &       & None  & 1.46$\scriptscriptstyle\pm$\scriptsize0.11 & \textbf{1.27}$\scriptscriptstyle\pm$\scriptsize0.01 & 1.31$\scriptscriptstyle\pm$\scriptsize0.03 \\
              &       & Explanatory variables & 1.33$\scriptscriptstyle\pm$\scriptsize0.04 & 1.29$\scriptscriptstyle\pm$\scriptsize0.04 & 1.30$\scriptscriptstyle\pm$\scriptsize0.03 \\
        1 & \acs{nRMSE}     & Drop  & \textbf{1.31}$\scriptscriptstyle\pm$\scriptsize0.08 & 1.29$\scriptscriptstyle\pm$\scriptsize0.01 & \textbf{1.26}$\scriptscriptstyle\pm$\scriptsize0.02 \\
              &       & Imputation & 1.50$\scriptscriptstyle\pm$\scriptsize0.16 & 1.40$\scriptscriptstyle\pm$\scriptsize0.06 & 1.32$\scriptscriptstyle\pm$\scriptsize0.08 \\
              &       & Drop-imputation & 1.62$\scriptscriptstyle\pm$\scriptsize0.11 & 1.32$\scriptscriptstyle\pm$\scriptsize0.01 & 1.31$\scriptscriptstyle\pm$\scriptsize0.06 \\
        \midrule
              &       & None  & 1.35$\scriptscriptstyle\pm$\scriptsize0.13 & 1.05$\scriptscriptstyle\pm$\scriptsize0.03 & 1.11$\scriptscriptstyle\pm$\scriptsize0.02 \\
              &       & Explanatory variables & 1.49$\scriptscriptstyle\pm$\scriptsize0.03 & 0.99$\scriptscriptstyle\pm$\scriptsize0.02 & 1.15$\scriptscriptstyle\pm$\scriptsize0.02 \\
        2 & \acs{nRMSE}     & Drop  & 1.23$\scriptscriptstyle\pm$\scriptsize0.07 & 1.01$\scriptscriptstyle\pm$\scriptsize0.02 & 1.07$\scriptscriptstyle\pm$\scriptsize0.02 \\
              &       & Imputation & \textbf{1.06}$\scriptscriptstyle\pm$\scriptsize0.03 & 0.99$\scriptscriptstyle\pm$\scriptsize0.01 & \textbf{0.98}$\scriptscriptstyle\pm$\scriptsize0.01 \\
              &       & Drop-imputation & 1.22$\scriptscriptstyle\pm$\scriptsize0.05 & \textbf{0.97}$\scriptscriptstyle\pm$\scriptsize0.02 & 0.99$\scriptscriptstyle\pm$\scriptsize0.03 \\
        \bottomrule
        \end{tabular}%
    }
\end{table}

\autoref{tab:autowp_shutdown-handling_disregard} shows the impact of shutdown handling methods on the forecasting error of the state\-/of\-/the\-/art autoregressive \ac{DL} methods (\ac{DeepAR}, \ac{NHiTS}, and \ac{TFT}) when \textit{disregarding shutdowns}.
With regard to \ac{WP} turbine no. 1, the shutdown handling method \textit{drop} results in the lowest forecasting error for \ac{DeepAR} and \ac{TFT}, while \textit{none} shutdown handling performs best for \ac{NHiTS}.
With regard to \ac{WP} turbine no. 2, the shutdown handling method \textit{imputation} achieves the lowest forecasting error for \ac{DeepAR} and \ac{TFT}, while \textit{drop\-/imputation} achieves the lowest errors for \ac{NHiTS}.
Interestingly, shutdown imputation in the past\-/horizon with \ac{OEM} \ac{WP} curve values only lead to improvement for \ac{WP} turbine no. 2.
This is because the \ac{OEM} \ac{WP} curve's forecasting error is comparably high for \ac{WP} turbine no. 1, as evident in \autoref{tab:autowp_benchmarks_disregard}.

\paragraph*{\textbf{Comparison of autoregressive and WP curve-based methods}}

\autoref{tab:autowp_benchmarks_disregard} compares the forecasting error of the autoregressive \ac{DL} methods \ac{DeepAR}, \ac{NHiTS} and \ac{TFT} (using the respective best\-/performing shutdown handling method) with \ac{WP} curve modeling-based forecasting methods.
Notably, even with the optimal shutdown handling method, the autoregressive \ac{DL} methods underperform compared to the methods based on \ac{WP} curve modeling.
Within the \ac{WP} curve modeling methods, AutoWP, \ac{MLP}, and \ac{SVR} achieve similar forecasting errors.
While AutoWP achieves the lowest \ac{nMAE} for both turbines, the \ac{SVR} achieves the lowest \ac{nRMSE} for \ac{WP} turbine no. 1, and the \ac{MLP} for no. 2, see \autoref{tab:autowp_benchmarks_disregard}.\footnote{
    The \ac{nMAE} measures the average magnitude of the errors, while \ac{nRMSE} gives a higher weight to larger errors, making it more sensitive to outliers.
}

\begin{table*}[]
    \centering
    \caption{%\mathversion{lm}
        Comparison of autoregressive \acs{DL} forecasting methods (\acs{DeepAR}, \acs{NHiTS}, and \acs{TFT}) to methods based on \acs{WP} curve modeling (\acs{OEM} curve, AutoWP, \acs{MLP}, \acs{SVR}, and \acs{XGB}) in terms of the test \acs{nMAE} when \textit{disregarding} shutdowns; results from extended evaluation of the conference paper \cite{Meisenbacher2024d}.
    }
    \label{tab:autowp_benchmarks_disregard}
    \resizebox{\textwidth}{!}{%
        % Table generated by Excel2LaTeX from sheet 'Evaluation 2'
        \begin{tabular}{cc|ccc|ccccc}
        \toprule
        \textbf{Turbine no.} & \textbf{Error} & \textbf{\acs{DeepAR}} & \textbf{\acs{NHiTS}} & \textbf{\acs{TFT}} & \textbf{\acs{OEM} curve} & \textbf{AutoWP} & \textbf{\acs{MLP}} & \textbf{\acs{SVR}} & \textbf{\acs{XGB}} \\
        \midrule
        1     & \multirow{2}[2]{*}{\acs{nMAE}} & 0.81$\scriptscriptstyle\pm$\scriptsize0.05 & 0.73$\scriptscriptstyle\pm$\scriptsize0.00 & 0.74$\scriptscriptstyle\pm$\scriptsize0.01 & 0.94$\scriptscriptstyle\pm$\scriptsize0.00 & \textbf{0.69}$\scriptscriptstyle\pm$\scriptsize0.00 & 0.83$\scriptscriptstyle\pm$\scriptsize0.01 & 0.70$\scriptscriptstyle\pm$\scriptsize0.00 & 0.84$\scriptscriptstyle\pm$\scriptsize0.00 \\
        2     &       & 0.77$\scriptscriptstyle\pm$\scriptsize0.02 & 0.66$\scriptscriptstyle\pm$\scriptsize0.02 & 0.67$\scriptscriptstyle\pm$\scriptsize0.01 & 0.67$\scriptscriptstyle\pm$\scriptsize0.00 & \textbf{0.61}$\scriptscriptstyle\pm$\scriptsize0.00 & 0.69$\scriptscriptstyle\pm$\scriptsize0.01 & 0.64$\scriptscriptstyle\pm$\scriptsize0.00 & 0.74$\scriptscriptstyle\pm$\scriptsize0.00 \\
        \midrule\\[-13pt]\midrule
        1     & \multirow{2}[2]{*}{\acs{nRMSE}} & 1.31$\scriptscriptstyle\pm$\scriptsize0.08 & 1.27$\scriptscriptstyle\pm$\scriptsize0.01 & 1.26$\scriptscriptstyle\pm$\scriptsize0.02 & 1.57$\scriptscriptstyle\pm$\scriptsize0.00 & 1.20$\scriptscriptstyle\pm$\scriptsize0.00 & 1.15$\scriptscriptstyle\pm$\scriptsize0.01 & \textbf{1.11}$\scriptscriptstyle\pm$\scriptsize0.00 & 1.22$\scriptscriptstyle\pm$\scriptsize0.00 \\
        2     &       & 1.06$\scriptscriptstyle\pm$\scriptsize0.03 & 0.97$\scriptscriptstyle\pm$\scriptsize0.02 & 0.98$\scriptscriptstyle\pm$\scriptsize0.01 & 1.07$\scriptscriptstyle\pm$\scriptsize0.00 & 0.93$\scriptscriptstyle\pm$\scriptsize0.00 & \textbf{0.92}$\scriptscriptstyle\pm$\scriptsize0.00 & 0.93$\scriptscriptstyle\pm$\scriptsize0.00 & 1.01$\scriptscriptstyle\pm$\scriptsize0.00 \\
        \bottomrule
        \end{tabular}%
    }
\end{table*}

\subsection{Insights}
\begin{figure*}
    \centering
    \includegraphics{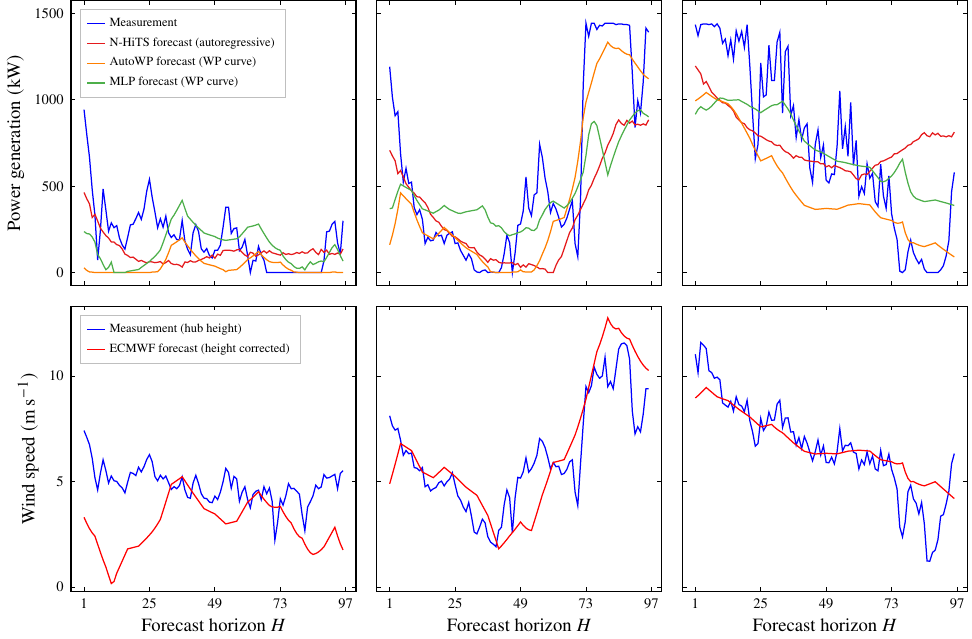}
    \caption{%\mathversion{lm}
        Exemplary comparison of \acs{WP} forecasts for \acs{WP} turbine no. 2 using \acs{NHiTS} (autoregressive), AutoWP and \acs{MLP} (\acs{WP} curve modeling) on days without shutdowns.
        While methods based on \acs{WP} curve modeling are only based on weather forecasts (\acs{ECMWF}), autoregressive methods also consider past \acs{WP} generation measurements when making forecasts.
        Legend of first column of a row applies to all columns in the row.
    }
    \label{fig:autowp_comparison-benchmarks_exemplary_line-plot}
\end{figure*}

\begin{figure*}
    \centering
    \includegraphics{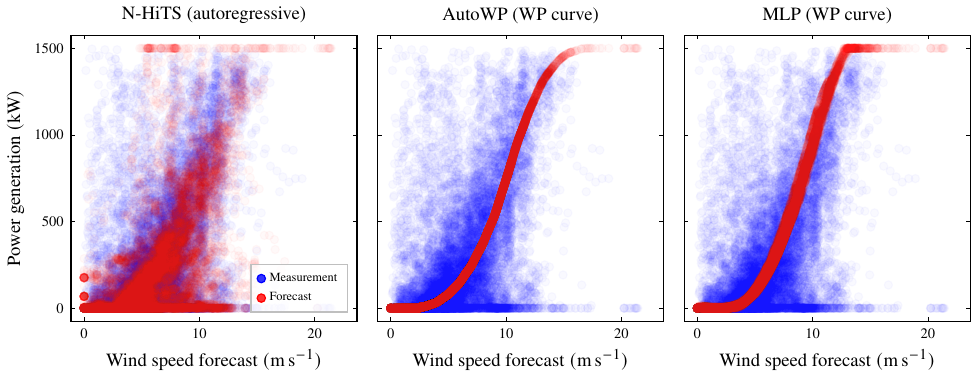}
    \vspace{-5mm}
    \caption{%\mathversion{lm}
        Relationship between wind speed forecast (\acs{ECMWF}) and \acs{WP} generation forecasts (\acs{NHiTS}, AutoWP, \acs{MLP}) for \acs{WP} turbine no. 2 when \textit{disregarding} shutdowns.
        AutoWP relies only on the height\-/corrected wind speed forecast, the \acs{MLP} is trained using the forecasts of wind speed and direction, temperature and air pressure, and \acs{NHiTS} additionally considers past \acs{WP} generation measurements.
        % The autoregressive method NHiTS considers past WP generation measurements alongside the weather forecast to make WP generation forecasts.
        % Contrarily, AutoWP relies only on the height\-/corrected wind speed forecast, and the MLP is trained using the forecasts of wind speed and direction, temperature and air pressure.
        No \acs{WP} generation despite moderate and high wind speeds are not shutdowns (already filtered out) but result from the error in the weather forecast, \ie, the wind speed forecast is above the cut\-/in speed while the realized wind speed is underneath.
        Legend of first column applies to all columns.
    }
    \label{fig:autowp_comparison-benchmarks_wpc_scatter-plot}
\end{figure*}

In the following, the specific insights of the evaluation are given.
\autoref{fig:autowp_comparison-benchmarks_exemplary_line-plot} provides an exemplary visual comparison of \ac{WP} generation forecasts for \ac{WP} turbine no. 2.
Apart from the \ac{WP} generation, also the wind speed measurement at hub height and the height\-/corrected wind speed forecast from \ac{ECMWF} is shown.
This exemplary comparison reveals that the wind speed forecast has a different influence on the resulting \ac{WP} generation forecast from \ac{NHiTS}, AutoWP, and \ac{MLP}.
Unlike AutoWP, which relies only on the height\-/corrected wind speed forecast, the \ac{MLP} additionally takes the wind direction, air temperature, and atmospheric pressure into account, and \ac{NHiTS} furthermore considers past \ac{WP} generation measurements.
\autoref{fig:autowp_comparison-benchmarks_wpc_scatter-plot} illustrates this in greater detail, plotting the \ac{WP} generation measurement $y$ and forecast $\hat{y}$ over the wind speed forecast $\hat{v}$.
The figure shows that \ac{NHiTS} attempts to compensate for the wind speed forecast's errors using the autoregressive part of the model.
However, this approach does not lead to significant error reduction in the evaluation data set.

\section{Discussion}
\label{sec:discussion}

In the following, the results of the two evaluation scenarios are discussed, the study's limitations are summarized, and recommendations are derived.

\paragraph*{Shutdown handling for methods based on autoregression}
With regard to the first scenario, we observe that shutdown handling methods significantly impact the forecasting error.
However, none of them consistently performs best across the autoregressive \ac{DL} methods \ac{DeepAR}, \ac{NHiTS}, and \ac{TFT}.
The already high computational training effort for these forecasting methods is further compounded by the need to identifying the best shutdown handling techniques, making model design even more resource\-/intensive.
This limits the scalable use and model deployment to a large number of individual \ac{WP} turbines.

\paragraph*{Comparison of autoregressive and WP curve-based methods}
With regard to the second scenario, the results show that even with the best\-/performing shutdown handling method, the considered autoregressive \ac{DL} methods do not improve over \ac{WP} curve modeling methods on the given data set.
Importantly, \ac{WP} curve modeling methods are static, \ie, only use day\-/ahead weather forecasts and thus cannot forecast regular shutdowns.
Such regular shutdowns can be considered either by having \ac{PK} of timed shutdowns or by designing a separate classification model that estimates whether the \ac{WP} turbine is in operation or shut down.
However, such a modeling approach requires a data set in which regular and irregular shutdowns are labeled.
The challenge of distinguishing between regular and irregular shutdowns becomes obvious in \autoref{fig:autowp_manual-shutdowns_heat-plot}, where recurring shutdown patterns are visible that are not strictly regular and may be interspersed with irregular shutdowns.

\paragraph*{Limitations}
Although the data of two \ac{WP} turbines with different shutdown patterns are considered, the data set is too small for drawing generalized conclusions.
Nevertheless, the evaluation reveals major issues for the scalable application of autoregressive \ac{DL} methods regarding redispatch planning.
For such forecasts, computationally efficient \ac{WP} curve modeling approaches are recommended.
Another limitation relates to the potentially untapped potential of \ac{HPO} to reduce forecasting errors.
However, even if \ac{HPO} could reduce forecasting errors of autoregressive \ac{DL} methods, it would further add to the computational effort required for the model design.
To improve the performance of the autoregressive \ac{DL} methods, labeled data sets would be required to distinguish between regular and irregular shutdowns.

\paragraph*{Benefits}
AutoWP can represent different site conditions characterized by terrain\-/induced air turbulence, leveraging an ensemble of different \ac{WP} curves.
More precisely, these site conditions influence the wind speed at which the \ac{WP} turbine reaches its peak power output.
These site conditions can differ greatly from the ones under which the \ac{OEM} \ac{WP} curve was determined, potentially resulting in a high forecasting error when using the \ac{OEM} \ac{WP} curve (as shown in the evaluation).
Although a site\-/specific \ac{WP} curve can also be considered when using \ac{ML} methods, AutoWP is advantageous because physical constraints in the \ac{WP} generation are implicitly considered, and only a few samples are required to train the model.
This scalability is crucial for the model deployment to a large number of distributed onshore \ac{WP} turbines to serve smart grid applications like redispatch planning.

\section{Conclusion and outlook}
\label{sec:conclusion_outlook}
Automated forecasting of locally distributed \ac{WP} generation is crucial for various smart grid applications in light of redispatch planning.
However, forecasting future shutdowns based on past redispatch interventions is evidently undesirable.
Therefore, this paper increases awareness regarding this issue by comparing state\-/of\-/the\-/art forecasting methods on data sets containing regular and irregular shutdowns.
We observe that autoregressive \ac{DL} methods require time\-/consuming data pre-processing for training and during operation to handle \ac{WP} turbine shutdowns.
In contrast, static \ac{WP} curve modeling methods that do not consider past values to make a forecast only require training data cleaning.
Within \ac{WP} curve modeling methods, AutoWP \cite{Meisenbacher2024d} is especially beneficial since it is computationally efficient, implicitly includes \ac{PK} about physical limitations in \ac{WP} generation, and achieves competitive performance.

Future work could address the stated limitations by extending the evaluation to a larger data set, labeled regarding regular and irregular \ac{WP} turbine shutdowns.
With such a data set, a separate classification model could be designed that predicts whether the \ac{WP} turbine is in operation or shut down.

\section*{Acknowledgments}
This project is funded by the Helmholtz Association under the Program ``Energy System Design'', the German Research Foundation (DFG) as part of the Research Training Group 2153 “Energy Status Data: Informatics Methods for its Collection, Analysis and Exploitation” and the Helmholtz Associations Initiative and Networking Fund through Helmholtz AI, and is supported by the HAICORE@KIT partition. Furthermore, the authors thank Stadtwerke Karlsruhe Netzservice GmbH (Karlsruhe, Germany) for the data required for this work.

\printbibliography

\printacronyms

\end{document}